\documentclass[11pt]{article}

\usepackage[preprint]{acl}
\usepackage{nert}
\usepackage{tcolorbox}

\usepackage{times}
\usepackage{latexsym}

\usepackage[T1]{fontenc}
\usepackage[utf8]{inputenc}

\usepackage{microtype}

\usepackage{inconsolata}

\usepackage{graphicx}

\usepackage{amsmath}
\usepackage{booktabs}

\usepackage[dvipsnames]{xcolor}

\newcommand{\an}[1]{\nertcomment{A}{A}{blue}{#1}}
\renewcommand{\finalversion}[1]{#1}

\newcommand*{\cat}[1]{{\textbf{\textsf{#1}}}}

\title{Speaking of Language: 
Reflections on Metalanguage Research in NLP}

\author{Nathan Schneider \\
  Georgetown University \\
  \eml{nathan.schneider@georgetown.edu} \\\And
  Antonios Anastasopoulos \\
  George Mason University \\
  \eml{antonis@gmu.edu} \\}

\begin{document}
\maketitle
\begin{abstract}
This work aims to shine a spotlight on the topic of metalanguage. 
We first define metalanguage, link it to NLP and LLMs, and then discuss our two labs' metalanguage-centered efforts.
Finally, we discuss four dimensions of metalanguage and metalinguistic tasks, offering a list of understudied future research directions. 
%We conclude that metalanguage is multi-faceted, and thus any (claims of) metalinguistic abilities of LLMs should be measured accordingly.
\end{abstract}

\section{Introduction}

%\nss{from my CAREER proposal:}

Language is so powerful that it can be reflected back on itself. All of the following English sentences expressly concern linguistic inventories, structures, and behaviors: 

\ex. People often confuse the ``George'' universities.

\ex. The expression ``kick the bucket'' is an idiom meaning ``die''.

\ex. Hebrew has a zero copula in the present tense.

\ex. Now read it in an Irish accent.

Sentences such as these may concern a particular instance of language use, or properties of a language or speaker in general; either way, they are \textbf{metalinguistic} in making linguistic phenomena (rather than the external world) the subject matter of a linguistic utterance.

%Below, the term \textbf{textual metalanguage} is used for natural language text in which natural language is also the subject matter.

% Automatic processing and understanding of textual metalanguage carries enormous potential for AI and NLP. 
% Imagine a machine translation system capable of reflecting on the grammatical and stylistic choices it is making, communicating these choices at a high level to a human user, and adapting to feedback from the user. 
% Or imagine giving high-level feedback to a summarization system (``please add some examples in the 2nd paragraph'').
% On the learning side, imagine that NLP models could seamlessly integrate explicit knowledge \textit{about} languages (in dictionaries, textbooks, reference grammars, and discussion forums) with distributional statistics from language \textit{use} in corpora.

%%%%%

%\nss{original version:}

% Meta-language refers to language used to talk about language itself. 
% Rather than describing the external world, metalanguage treats linguistic expressions, structures, and meaning as its subject matter.
% Examples include explaining why a sentence is grammatical, defining a word, or discussing how a phrase is interpreted. 

Any kind of formal notation that elucidates linguistic properties can also be considered metalanguage, e.g.:

\ex. One morning I shot [$_{\text{NP}}$ an elephant in my pajamas].

\ex. an\textsubscript{\texttt{DT}} elephant\textsubscript{\texttt{NN}} in\textsubscript{\texttt{IN}} my\textsubscript{\texttt{PRP\$}} pajamas\textsubscript{\texttt{NNS}}

\ex. \textit{shot}(\textit{I}, \textit{elephant}) $\wedge$ \textit{in}(\textit{elephant}, \textit{pajamas})

%Descriptions of language can be subdivided into \textbf{symbolic metalanguage} that uses formal notation, and \textbf{natural metalanguage}---natural language utterances discussing language (e.g., mentioning a word as a word rather than using it). 

Both kinds of metalanguage enable humans to reflect on linguistic form, meaning, and use, which is why metalanguage is central to fields such as linguistics, language pedagogy, rhetoric, and even law and policy.

The recent advent of fluent multipurpose chatbot tools powered by large language models (LLMs) puts a new focus on metalanguage---or at least we argue that it should.
Despite some research on metalanguage and on the metalinguistic abilities of LLMs, the topic remains an understudied one. 
%However, we argue its importance noting that human language competence fundamentally includes the ability to reason about language itself. 
Yet many real-world tasks in spheres such as language learning and law rely on metalinguistic reasoning rather than simple content understanding. 
Evaluating whether LLMs can process, generate, and learn from metalanguage therefore provides a crucial test of %how human-like their (linguistic?) capacities truly are. 
adaptability and robustness.
%Moreover, studying these abilities can reveal both strengths and limitations of current models, 
Zeroing in on the processing of metalanguage, we believe,
%may yield insights that improve accuracy, interpretability, and responsible deployment in high-stakes 
will ultimately serve applications in important domains where meaning and interpretation are central.

\begin{table*}\centering %\small
\begin{tabular}{@{}lcc@{}}\toprule
                    & \textbf{symbolic} & \textbf{natural} \\ \midrule
    \textbf{instance-level}  & tagging, parsing & answering a question about the grammaticality of a sentence \\ \midrule
    \textbf{system-level} & grammar rule induction & generating a dictionary definition for a word \\ \bottomrule
\end{tabular}
\caption{Examples of metalinguistic description tasks (where the metalanguage is language-system--oriented in nature). The metalinguistic inputs and/or outputs in question may be symbolic or natural, and can be formulated at the level of individual instances (tokens) or at the level of an entire system of language (generalizations).}
\label{tab:tasks}
\end{table*}

\section{Defining ``Metalanguage''}

The human ability to use language to communicate rests on knowledge that is mostly \emph{implicit}. Linguistics and other fields \textbf{communicate explicit conceptualizations of language phenomena} via metalanguage \citep{berry-05}. For example, the statement ``5-year-old children can productively form regular plurals of nouns'' will be incomprehensible to most adults, not to mention the 5-year-olds in question. Not only do the plural-producing 5-year-olds not know the \emph{word} ``noun'', they presumably have not learned any grammar to the point of comprehending the \emph{concept} of ``noun''. The metalanguage of a discipline such as linguistics thus reflects many of the concepts at the core of disciplinary expertise.

We provide brief terminological definitions below to setup the stage for the rest of the paper.
\textbf{Natural metalanguage} is text in a natural language that is interpreted to be about language, grounded in particular utterances or general behavior by a speaker or language community.
\textbf{Symbolic metalanguage} is formal notation that encodes aspects of language (or instances of language use) in a way that explicitly surfaces relationships and patterns within the language system.\finalversion{\footnote{We focus here on metalanguage that is expressed in human-understandable formats, so we will not discuss ``style vectors'' or ``task embeddings'' as potential metalanguage.}}
In general, \textit{a} symbolic metalanguage is a formal system or controlled vocabulary for describing a phenomenon.
Finally, we note that quotations (a reference to another speech act or text) are a specific instance of metalanguage that serves (usually) a more narrow/specific communicative purpose.

Broadly, a \textbf{metalinguistic task} is one that necessarily processes, leverages, produces, or is defined with metalanguage.
Below we discuss kinds of metalinguistic tasks in the context of LLMs (\cref{sec:llms}).

\section{Why Study Metalanguage (in NLP)?}\label{sec:why}

Metalinguistic inquiry of one sort or another is common in a wide range of fields and applications. 
Linguistics, of course, is entirely about the study of language. 
Consider also: language teaching and learning (e.g., second language learners asking for advice about how to use a word or construction); lexicography; 
literary studies; 
and law (the interpretation of legal rules).
In these domains, amateur or professional language analysts \textit{consume} instances of language use (in some cases using highly customized corpus search tools), and/or \textit{produce} large quantities of textual metalanguage in textbooks, dictionaries, online discussion forums, legal opinions, and scholarly publications. 
One impetus, then, for NLP study of metalanguage is to develop tools for metalinguistic inquiry.

Another motivation comes from the inherent goals of modeling language and linguistic meaning. For humans engaged in scientific work, natural as well as formal languages are indispensable when developing theories and making predictions. Within NLP, the tradition of analyzing linguistic grammar and meaning with symbolic structures is one incarnation of metalinguistic NLP \citep{opitz-25}. (Thus, the study of syntactic parsing, for example, is inherently metalinguistic.)

\subsection{Metalanguage and LLMs}\label{sec:llms}

In light of the current fascination with large language models (LLMs), it is worth breaking down where metalanguage may come into play in this paradigm, and what studies have or might shed light on its role.
%\subsection{Metalinguistic Modes}

The phenomenon of metalanguage confronts different forms of interaction with an LLM system. We outline these metalinguistic modes below.

\paragraph{Metalinguistic instructions.} 
If a chatbot user explicitly requests a piece of writing---whether it is a summary, translation, homemade pizza recipe, or humorous limerick about cheese---that user is speaking metalinguistically.\footnote{We consider an instruction metalinguistic if it makes any reference to communication or the linguistic nature of the input or output. Thus ``Tell me a joke.''\ and ``limerick about cheese'' are metalinguistic; ``What is the capital of France?'' is~not.} Thus the practices of instruction tuning and prompting are tied up with metalanguage to some extent.
This implies that the presence (or not) and the extent of metalinguistic intent should perhaps inform the evaluation of LLMs in general: do they perform better or worse when the instructions are metalinguistic or not?

\paragraph{Metalinguistic description tasks.} 
By this we mean tasks that center \textit{systematic} aspects of language (or a language).
Requesting a definition, grammatical analysis, or explanation of meaning all presuppose a set of conventions constituting a linguistic system, and seek a description that somehow unpacks the conventions or how they apply to a particular instance.
\Cref{tab:tasks} illustrates tasks that involve \textit{symbolic} or \textit{natural} metalanguage describing linguistic \textit{instances} or \textit{generalizations}.
Not all language-manipulation tasks qualify here: machine translation of a sentence, for example, does not in itself reference the organization of either language system (though a translation may be part of a larger explanation of linguistic patterns constituting a metalinguistic description).\footnote{There is probably no bright line that demarcates this category. Between ``Proofread this paragraph'' and ``Indicate the grammatical errors in this paragraph'', the latter more overtly invokes a goal of linguistic system--based description, but in practice these serve very similar user needs. An alternative definition going beyond our focus could take into account user intent so that e.g., producing a free translation for an interlinear gloss of an example in a reference grammar could qualify as such a metalinguistic description task.}
%\an{The question of user(?) intent also is of relevance: producing a free translation to be used as part of an interlinear gloss of an example in a reference grammar, definitely has metalinguistic intent and could thus still qualify as such. But here we primarily focus on tasks which necessarily require some sort of ...? }\nss{seems similar to footnote 2. can it be expressed there?}
%\finalversion{\nss{Dawson: cite Hu/Levy here?}}

\paragraph{Metalinguistic interpretation and explanation.}
To better understand how ``black box'' models of language operate, one route is to look for correlations between representations or behaviors in the model and their counterparts as described metalinguistically for human language. For example, attempts have been made to localize grammatical knowledge amongst a tangled web of neural network components \citep[e.g.,][]{liu-19,tenney-19,aoyama-22,wang-22}. Other work has investigated model behavior vis-\`{a}-vis its generations or probability distributions \citep[e.g.,][]{blimp,hu-23}.
We return to metalinguistic interpretability in \cref{sec:directions}.

\section{A Tale of Two Labs}
\label{sec:labs}

Research programs within the authors' research groups have prioritized metalinguistic NLP.\footnote{Supported in part by NSF awards ``CAREER: Metalinguistic Natural Language Understanding'' (Schneider) and ``CAREER: Leveraging Grammar Books to Develop Language Technologies for Data-Scarce Languages'' (Anastasopoulos).}
We give an overview of several such efforts:
\begin{itemize}
\item in Anastasopoulos's lab at George Mason University, studies featuring documentary linguistics and low-resource NLP (\cref{sec:grammars,sec:patterns});
\item in Schneider's lab at Georgetown University, studies motivated by second language learning and legal interpretation (\cref{sec:learning,sec:law}).  %\nss{could also talk about work by Shira/Jakob with semantic structures, but that is a bit older}
\end{itemize}
The concluding sections will discuss emerging themes and dichotomies.

\subsection{Learning from Reference Grammars}\label{sec:grammars}

The idea of learning using already-defined grammars is not new; it is in fact one of the first ideas tried out in the early era of symbolic NLP. %~\an{oooof citations?}. 
However, using the text of a reference grammar \textit{as is} to facilitate the creation of language technologies for a given language has only recently come within reach, due to LLMs' capabilities.

Calls to ``mobilize the archive'', in particular for data-scarce languages, aim to encourage research that leverages linguistic documentation efforts~\cite{bird-2022-local}.
In seminal work, \citet{tanzer2023mtob} did exactly that, incorporating dictionaries, sentences, and grammar books to perform machine translation using LLMs in a zero-shot setting, i.e., in a language without \textit{any} other data available
(``Machine Translation from One Book'').
This is perhaps akin to how a documentary linguist or any second-language learner could potentially learn a new language (at least if they did not have access to a teacher or said language's speakers).

In follow-up work, \citet{hus-anastasopoulos-2024-back} explored this grammar-based paradigm on 16 languages. 
\iffalse
For translating from an entirely unseen language, we crafted prompts $\pi(\mathbf{x}, t, d, s, g)$ that additionally included: 
    (1) word-level translations $d$ obtained from a bilingual dictionary $\mathcal{D}$; %, selected for their similarity to the words of the given source sentence;
    (2) a few parallel sentence examples $s$; and %, selected from a small collection of parallel sentences $\mathcal{S}$ for their similarity to the given source sentence; and
    (3) excerpts $g$ from a grammar book $\mathcal{G}$. %, also selected for similarity to the source sentence using longest common substring distance.
\fi
Our initial findings were particularly encouraging. For translating extremely low-resource languages like Chuvash, Dogri, and Kalamang into English, providing a combination of dictionary entries and the full grammar book yields almost usable translations (with chrF++ scores between 25--55).
Other ongoing work attempts to integrate such approaches into a documentary linguist's workflow---in this case, working on Nepal's Kulung languages~\cite{taguchi2025digital}.

However, concurrent and followup work has called into question whether the current generation of LLMs can trully understand and leverage metalinguistic content in the form of a reference grammar~\cite{aycock-etal-2025-iclr,marmonier-etal-2025-explicit}. 
Regardless, we believe that the potential for reducing data requirements for under-resourced languages of already extremely under-served communities makes this a worthy research direction.

%\finalversion{Add paragraph about OCR efforts over reference materials}

\subsection{Inducing and Describing Patterns in Data}\label{sec:patterns}

Another line of work aims at \textit{generating} metalanguage (symbolic or natural). In particular, we aim at simulating the work of a linguist or a language teacher, producing output that describes a language system, based on raw text samples.

\begin{figure*}[t]
    \centering
    \includegraphics[width=.9\linewidth]{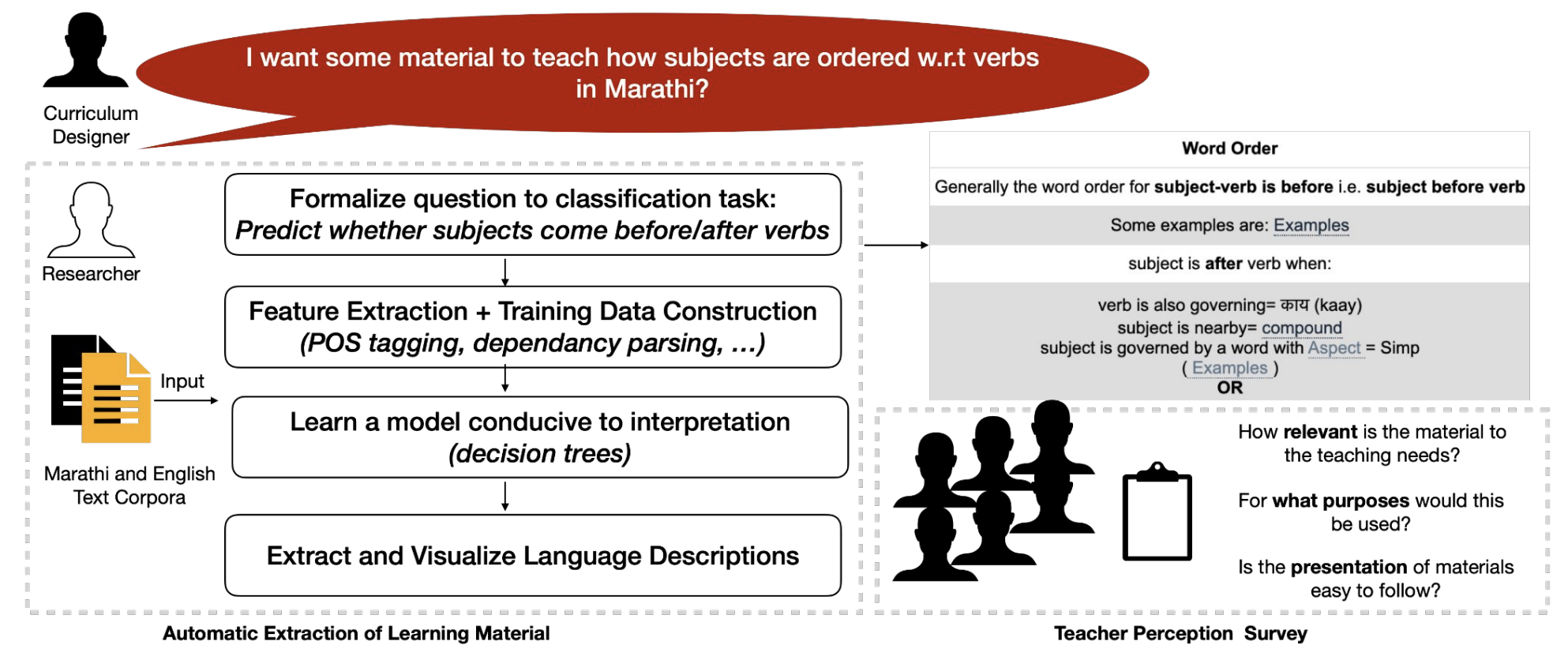}
    \caption{Workflow for the collaboration of NLP researchers and language-learning curriculum designers, to create pedagogical materials~\cite{chaudhary-etal-2023-teacher}. The input and intermediate and final outputs include metalanguage.}
    \label{fig:aditi}
\end{figure*}

The notion of describing a language “in its own
terms” based solely on raw data has an established
tradition in descriptive linguistics~\cite{harris1951methods}.
Early work included discovering morphosyntactic agreement~\cite{chaudhary-20} or lexical selection preferences~\cite{chaudhary-etal-2021-wall}, tying it also to educational applications, by presenting these rules along with selected examples to be used by L2 teachers~\cite{chaudhary-etal-2023-teacher} -- see an illustration in Figure~\ref{fig:aditi}.
Earlier work by~\citet{howell2017inferring} aimed to predict the case systems of endangered languages and \citet{zamaraeva2016inferring}
inferred morphotactics from IGT using $k$-means
clustering.

The above line of work is somewhat orthogonal to the works that have typologists as their target audience. 
Linguists and researchers have long undertaken initiatives to collect linguistic properties in machine-readable formats.
\textit{WALS} \cite{wals} is one such example which can tell us, for instance, that English objects occur after verbs, or that Turkish pronouns have symmetrical case. 
\textit{Grambank}~\cite{grambank} is the latest typological database: it covers 2,467 language varieties, capturing a wide range of grammatical phenomena in 195 features, from word order to verbal tense %, nominal plurals, 
and many other well-studied comparative linguistic variables.
Most works on the NLP side aim to fill in the missing entries in such databases, producing a structured typological description of a language based on raw text samples~\citep[][\textit{inter alia}]{daume-07,bjerva-etal-2020-sigtyp}. See~\citet{baylor-etal-2023-past} for additional discussion on the usefulness of such work for NLP in general. 
More recently, \citet{arcon-26} converted each entry of the WALS database into a question-answer pair, in order to test LLMs' typological knowledge.
\citet{hus-anastasopoulos-2026-sigtyp} pose similar typological questions but they also assume access to reference grammars for the languages in question.

Notably, to our knowledge, there is no substantial progress in integrating LLMs with field linguists' or typologists' workflow. Most works listed above are from the pre-LLM era.
One exception is the rather exciting exploration of the metalinguistic reasoning capabilities of LLMs focused on small, artificial problems inspired (or directly taken) from linguistic olympiads, such as the PuzzLing Machines~\cite{sahin-20} and LingOly~\cite{bean2024lingoly} benchmarks (see also \cref{sec:directions}).
But all such problems with their guarantees of a single correct solution and carefully curated data to reach it barely mimic real-world settings, where incomplete and ambiguous data render the task significantly harder.
The very recent work of \citet{yang2025linggym} that tests whether LLMs can be used to gloss unknown lexical items is one step in the above direction.\footnote{We note, although, that it still operates in a less noisy environment and with more available information as input than one might expect in real-world settings.}

%(incl. work with learners as the target audience)

\begin{figure*}
    \centering
    \begin{subfigure}[b]{0.53\textwidth}
        \centering
        \includegraphics[width=1\textwidth ,trim = 0cm 5cm 9.25cm 4cm, clip]{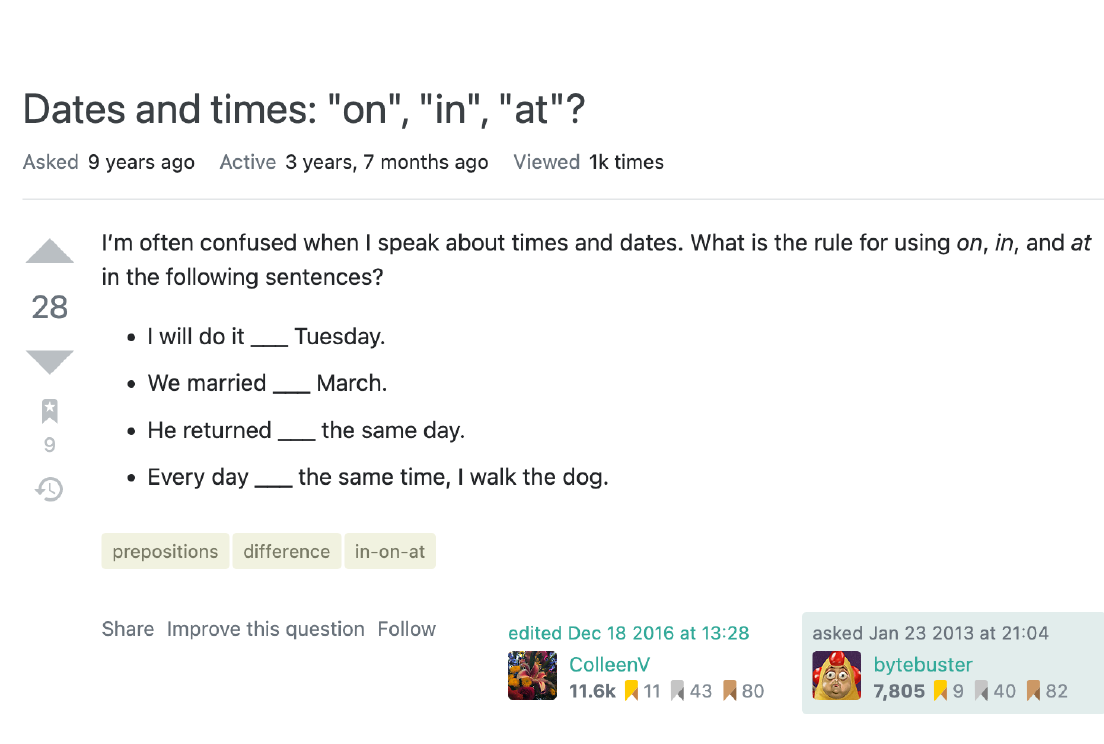}
        \caption{Question}
        \label{fig:q}
    \end{subfigure}
    \hfill
    %\vspace{0.1cm}
    \begin{subfigure}[b]{0.46\textwidth}
        \centering
        \includegraphics[width=1\textwidth , trim = 0cm 3.5cm 4cm 0cm, clip]{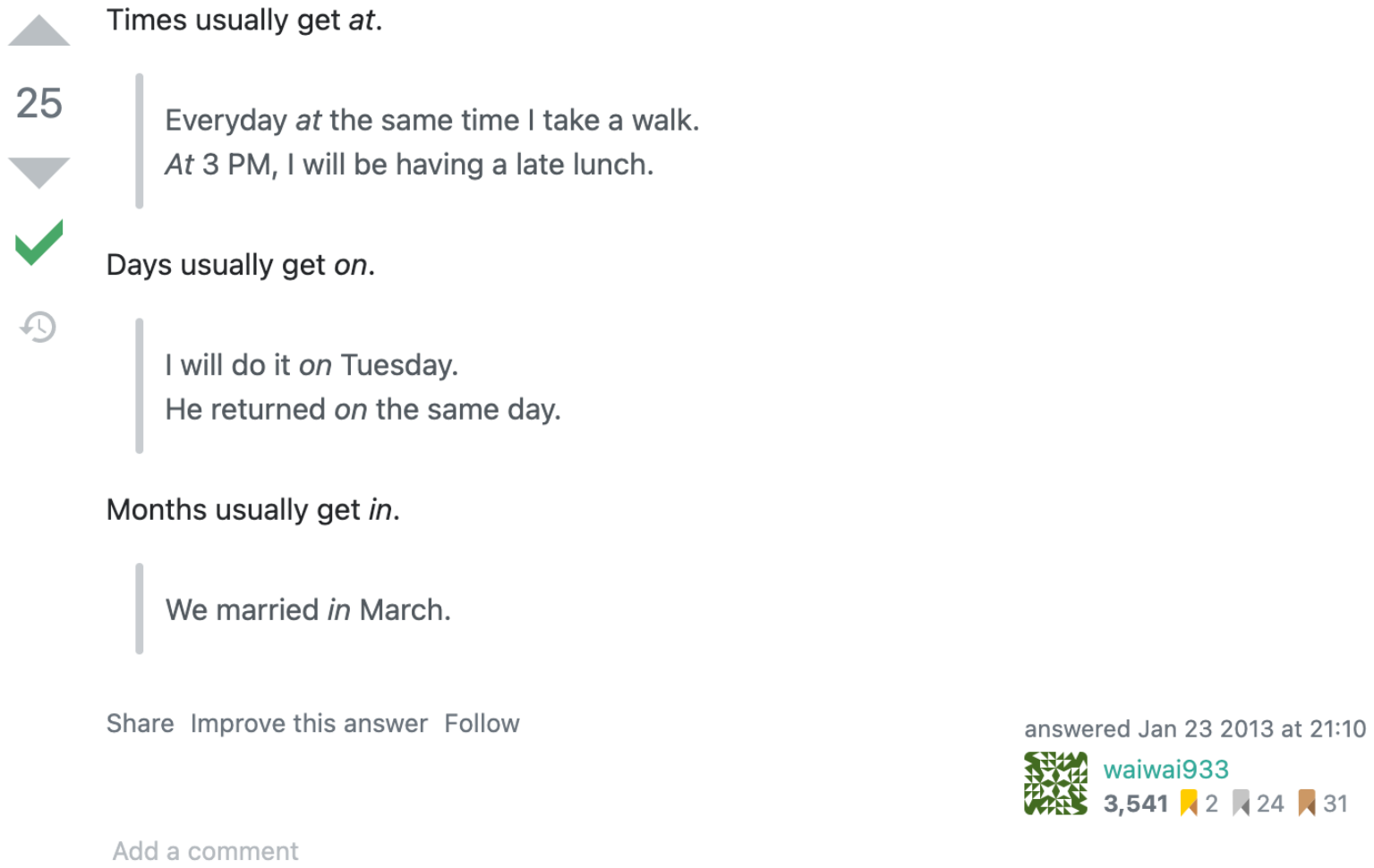}
        \caption{Answer}
        \label{fig:a}
    \end{subfigure}
        %\vspace{0.1cm}
        \caption{Screenshots of a page on the English Language Learner Stack Exchange site, which is included in the ELQA dataset \citep[from][]{behzad-23}. The source page is \url{https://ell.stackexchange.com/questions/12/dates-and-times-on-in-at}.}
        \label{fig:ell-stackexchange}
        \vspace{-1em}
\end{figure*}

\subsection{Language Learning Domain}\label{sec:learning}

An important domain for metalanguage is the realm of language teaching and learning, whether in a classroom or in a less formal context such as an online forum or chatbot interaction.
Complementing the extraction of symbolic rules for educational purposes described above (\cref{sec:patterns}), it is valuable to investigate \textbf{natural} metalanguage scenarios in the language learning domain.

Here we highlight two studies of LLM processing of natural metalanguage (NML). 
\Citet{behzad-23} collected data from two online English discussion forums (one designed specifically for L2 English learners) in order to construct a benchmark for English Language Question Answering (ELQA). The metalinguistic questions in this dataset span a range of topics including vocabulary, grammar, and meaning; some, for example, inquire about sentence grammaticality, while others request help with expression or passage interpretation, or asked general questions about linguistic conventions. An example of a question and corresponding answer, both user-generated, appear in \cref{fig:ell-stackexchange}: the question/answer pair mix a system-level overarching issue (about the rules for prepositions in times and dates) with specific exemplar instances.

Sampling questions and answers from ELQA, \citet{behzad-23} conducted a human evaluation pitting user responses against responses from LLMs (including GPT-3 with few-shot learning or finetuning). Note that this kind of question answering is an NML-to-NML task. Overall, the best GPT-3 setup was highly fluent across the board, and gave accurate answers for many of the questions, but in some cases underperformed the highest-rated user answer for accuracy. This suggests that LLMs may be helpful interlocutors in answering learners' metalinguistic questions, but at times will make mistakes (with the caveat that today's state-of-the-art models have not yet been evaluated).

In a followup study, \citet{behzad-etal-2024-ask} addressed the \emph{crosslinguistic} dimension of learner QA, speculating that L2 learners using an LLM may frame questions in their native language. 
This raises the question of whether the pairing of prompt-language and target language matters.
This study used a controlled paradigm of minimal pair grammaticality judgment\footnote{Other work has studied metalinguistic prompting for grammaticality judgments \citep[e.g.,][]{hu-23}.} (given an original and a corrected sentence from a grammatical error correction dataset) so that it would be possible to independently manipulate the language of the prompt template and the language of the target sentence. 
The languages tested were English, German, Russian, Ukrainian, and Korean.
The tested models displayed a great deal of sensitivity to the choice of prompt-language, suggesting that the apparent multilinguality of an LLM does not guarantee stable metalinguistic behavior across languages.

\subsection{Legal Interpretation}\label{sec:law}

First, a bit of background. \textbf{Legal interpretation} is the enterprise of determining the meaning of a rule expressed in natural language \citep{brannon-23-interp}. This frequently arises in judicial cases, where a judge must determine the extent of a category in order to decide whether it applies to the facts of the case. (If the rule is ``no vehicles are allowed in the park'', and ``vehicle'' is not specifically defined, should it be read to encompass skateboards? Wheelchairs? Ambulances? Are there contextual clues that shed light on the scope of the category?) The vehicle rule is a hypothetical example, but real cases similarly discuss the semantic interpretation of a term, such as the ``landscaping'' controversy summarized in \cref{fig:landscaping-example}. Other cases implicate grammatical ambiguities in the wording of a statute or contract.
Many U.S.\ judges subscribe to the philosophy that the starting point for interpreting legal language is the general language---they seek to determine the ``ordinary meaning'' of the text per contemporaneous usage (as members of the public might interpret it today, or when the law was enacted).
This ``textualist'' perspective has engendered scholarly examinations of the nature of meaning in language, and the principles judges have articulated in an attempt to make language analysis rigorous and objective.
But the practice of textualism has come under fire from scholars who contest the supposed neutrality of these principles---suggesting that they are poorly formulated, rooted in misconceptions about linguistic meaning, or so malleable that they can be manipulated to support any outcome \citep{eskridge-23}.

Through collaborations with law professor Dr.~Kevin Tobia---who has advocated for empirical approaches like survey research to ascertain ordinary meaning \citep[e.g.,][]{tobia-20,tobia-22,waldon-25-artifact}---Dr.~Schneider and his lab are investigating computational linguistic and NLP tools for textualist inquiry. There are several threads of investigation.

\begin{figure}[t!]
\centering
\begin{tcolorbox}\sf
John is a contractor with insurance that covers property loss, damage, or personal injury claims that arise due to his `landscaping' work.
\\
\hspace*{1.5em}John is employed by a family, the Smiths, to install an in-ground trampoline in the family’s backyard. A few years after John completes the project, the Smiths successfully sue John for injuries that their daughter sustained while playing on the trampoline. John files a claim with his insurance company to recover losses incurred from the lawsuit.
    \end{tcolorbox}
    Considering just how ``landscaping'' would be understood by ordinary speakers of English, is John covered by the insurance—yes or no?
    \caption{A legal interpretation scenario represented as a QA task with binary questions. The example is based on the case \href{https://media.ca11.uscourts.gov/opinions/pub/files/202212581.pdf}{\emph{Snell v.~United Specialty Insurance Co.}}\ and constructed in the style of one of the prompting formats studied by \citet{purushothama-facct}.}
    \label{fig:landscaping-example}
    % \vspace{-10pt}
\end{figure}

\paragraph{Are LLMs trustworthy as tools for answering interpretive questions?} 
Already judges have begun to entertain the possibility that difficult interpretive questions might be outsourced to LLM chatbots, on the rationale that huge amounts of ordinary usage in training data would entail accurate (perhaps superhuman) metalinguistic conclusions about meaning.\footnote{One judge writes: ``models train on a mind-bogglingly enormous amount of raw data [across many genres]. Because they cast their nets so widely, LLMs can provide useful statistical predictions about how, in the main, ordinary people ordinarily use words and phrases in ordinary life'' \citep{snell}. This fails to appreciate the distinction between learning implicit usage patterns, and being able to articulate those patterns metalinguistically in response to a prompt.}
\Citet{waldon-25-llms} push back against this assumption, attributing it to myths about how LLMs work. Further experimentation by \citet{purushothama-facct} and \citet{petersen-26} examines prompt sensitivity and alignment with human consensus. (One of the tested prompt formats appears in \cref{fig:landscaping-example}.) The results so far indicate that state-of-the-art platform models are less sensitive to prompt framing than smaller-scale models, and achieve some level of correlation with human judgments, but are not immune to giving an implausible answer when asked for a binary judgment. They are therefore \emph{not} a silver bullet for resolving difficult questions. If they have any utility for interpretive reasoning, it is probably as a brainstorming aid: the system can be asked to generate arguments for and against a position, provided the human judge critically evaluates all claims \citep{waldon-25-llms}.

\begin{table*}
    \centering\smaller[1]
    \begin{tabular}{lp{10.5cm}}
        \toprule
        \textbf{Category} & \textbf{Definition} \\
        \midrule
        \cat{Focal Term} (FT) & Word or phrase used metalinguistically and/or whose meaning is under discussion.\\
        \midrule
        \cat{Definition} (D) & Succinct, reasonably self-contained description of what a word or phrase means. Need not be exhaustive. May also be negative—defining a word by what it’s not.\\
        \midrule
        \cat{Metalinguistic Cue} (MC) & Word or short phrase cueing nearby metalanguage.\\
        \midrule
        \cat{Direct Quote} (DQ) & Span of text inside quotation marks.\\
        \midrule
        \cat{Legal Source} (LeS) & Citation or mention appealing to a legal document or authority.\\
        \midrule
        \cat{Language Source} (LaS) & Citation or mention appealing to an authority on language.\\
        \midrule
        \cat{Named Interpretive Rule} (NIR) & Mention of a well-established interpretive rule or test used to support an argument about the meaning of a word or phrase.\\
        \midrule
        \cat{Example Use} (ES) & Intuitive, quoted, or hypothetical examples that demonstrate a word/term can or cannot be used in a certain way.\\
        \midrule
        \cat{Appeal to Meaning} (ATM) & An explicit argument, implicit value judgment, or other statement indicating how one should go about interpreting meaning (e.g., by appealing to common sense, ordinary meaning, or the language of another statute).\\
        \bottomrule
    \end{tabular}
    \caption{Categories of metalanguage annotated in the CuRIAM corpus \citep[from][]{kranzlein-24}.}
    \label{tab:cats}
\end{table*}

\paragraph{How prevalent are different facets of metalanguage in judicial opinions?} \Citet[ch.~5]{kranzlein-24,kranzlein-diss} examined this as a computational social science question by taxonomizing and tagging different facets of metalanguage in a corpus of Supreme Court opinions.
An example of a metalinguistic sentence from one of these opinions \citep{breyer-23}:

\ex.\label{ex:breyer}First, the Act defines ``pollutant'' broadly, including in its definition, for example, any solid waste, incinerator residue, ``heat,'' ``discarded equipment,''' or sand (among many other things). §502(6), 86 Stat. 886.

Notably, sentence \cref{ex:breyer} includes several parts: \textbf{metalinguistic cue} words that denote linguistic units or processes (``defines'', ``definition''); a \textbf{focal term} being defined (``pollutant''); portions of a \textbf{definition} of the focal term; and a citation to a \textbf{legal source}.
\Citet{kranzlein-24} envision metalanguage category tagging as an information extraction task, and annotate these categories in the CuRIAM corpus. (The full list of categories appears in \cref{tab:cats}.)

\Citet[ch.~5]{kranzlein-diss} then trains a tagger for these categories: a \textbf{metalanguage identification} task. With automatic tagging, he conducts a content analysis of three decades of Supreme Court opinions. His analysis of metalanguage use over time points to an increase of some of the categories from 1986 to 2018, consistent with the growing popularity of textualism as an interpretive philosophy.

\paragraph{Do judicial canons of construction reflect accurate generalizations about linguistic usage?} 
Over time, textualist judges have developed a suite of heuristics known as \textbf{canons of construction}. These assert preferences for resolving certain ambiguities in the text \citep{scalia-12,brannon-25-canons}.\footnote{The Nearest-Reasonable-Referent Canon, for example, makes recommendations about how to disambiguate the syntactic attachment of a modifier \citep{scalia-12}.}
In response to calls for basing canons on stronger empirical foundations \citep{tobia-22-canons},
we have sought to critically examine the canons via computational linguistic techniques: namely corpus analysis (e.g., compiling judicial opinions referencing a particular canon in order to establish how it tends to be applied), treebanking \citep[e.g., to establish the most frequent resolution of syntactic ambiguity in statutory text;][]{waldon-25-legalcgel}, and semantic annotation \citep{wells-25-abstract}.
%\finalversion{\nss{cite work by Michaela and Lillian when it comes out}}
These investigations are ongoing. Our hope is that they will lead to a clearer articulation of the canons (with precise terminology from linguistics), as well as empirical data about the reliability of each canon in practice.

%- This work connecting linguistics and computation with law is having real impact. 1) VanDerStok, 2) sentencing commission, 3) insurance industry?
%(not sure if it's worth also discussing the legal-audience work like VanDerStok)

\section{Dichotomies in Metalanguage Research}
\label{sec:discussion}

%\an{Do we draw contrasts between our two labs?, e.g., Antonis focuses on using system-level metalanguage for downstream applications, Nathan focuses on instance-level questions for law; for learners, Antonis focuses on discovering and show-casing examples of grammar rules, Nathan on metalinguistic QA} \an{I don't feel strongly about doing that though, I don't think it's necessary}

Organizing metalanguage research, even when not taking a very broad view of metalanguage, requires considering multiple axes of analysis. We observe the following notable dichotomies (two of which are highlighted in \cref{tab:tasks}):
\begin{itemize}[leftmargin=*]
    \item \textit{system-level vs.~instance-level metalanguage:} system-level metalanguage targets general properties of a linguistic system (e.g., grammatical rules, constructions, or typological facts), at the level of the entire language or subpopulation of the language community; instance-level metalanguage concerns specific linguistic tokens or contexts (e.g., explaining why a given sentence is ambiguous).
    \item \textit{monolingual vs.~multilingual:} the necessary LLM capabilities as well as system requirements and design would likely need to differ for metalinguistic inquiries targeting a single language, contrasting a pair of languages, or if the focus is specifically on second-language settings.
    \item \textit{symbolic vs.~natural metalanguage:} employing formalized notations such as parse trees, symbolic metalanguage enables precision but is less accessible to lay users than using ordinary (natural) language to describe linguistic phenomena.
    \item \textit{processing vs.~generation of metalanguage:} while processing metalanguage as input might largely evaluate models' comprehension, generation reveals whether models externalize linguistic reasoning in useful ways. Of course, many applications such as legal analysis and educational assistance require both capabilities.
\end{itemize}

By necessity, any metalanguage-related work will occupy a position along many of these axes, as do many of our works discussed in \cref{sec:labs}.

\section{Research Directions}
\label{sec:directions}

Research connecting metalanguage and LLMs asks how well LLMs fare on different kinds of metalinguistic tasks; why; and to what end.
We list a few specific directions below.
Some of these have already been subjects of inquiry, for which we give illustrative citations from our labs and others.
Many directions though are, to the best of our knowledge, heretofore unexplored:\footnote{Here we include work with transformer models like BERT and GPT-2, though our main focus in this paper is on contemporary prompt-based LLMs.} 

\paragraph{Intrinsic evaluation questions}
Here we focus on research directions that aim to evaluate the metalinguistic capabilities of LLMs:
\begin{enumerate}
\item Can an LLM solve linguistic structure NLP tasks \citep[e.g.,][]{ettinger-23,tian-24}, analysis problems in theoretical linguistics \citep{begus-25}, or language puzzles \citep{rozner-21,sahin-20,bean2024lingoly,chi-24,sanchez-25,choudhary2025unveiling}?
\item How well can models understand self-referential language (``This sentence is short.'')? \citep{thrush-24}
\item How well can models distinguish mentions vs.~uses?\footnote{\emph{Mentioned language} when text refers explicitly to a linguistic entity like a word or sentence. A thorough definition is given by \citet[ch.~2]{wilson-11}, and a study of statistical classifiers is presented by \citet{wilson-13}.}  \citep{kranzlein-diss} (discussed in \cref{sec:law})
\item How does the choice of the (natural or formal) language in which metalanguage is formulated affect model behavior in relation to the described language? \citep{behzad-etal-2024-ask} (discussed in \cref{sec:learning})
\item Are LLMs sensitive to pragmatic phenomena like so-called \textit{metalinguistic negation} \citep{horn-85},\footnote{Also known as \emph{frame-rejecting negation}; an example is ``John isn't being thrifty, he's just downright stingy'' \citep[p.~243]{fillmore-85}.} where the speaker uses negation to signal disagreement with a choice of words, and quotation, where the speaker is not necessarily committing to the same perspective as the source they are quoting? \citep[focusing on detecting whether hate speech and misinformation reflects the speaker's perspective]{gligoric-24}
\end{enumerate}

%\finalversion{\nss{TODO: definitions e.g. https://arxiv.org/abs/2603.11687}}

\paragraph{Interpretability questions}
Next we outline inquiries that are paramount in order to understand \textit{why and how} metalinguistic abilities arise in LLMs:
\begin{enumerate}[resume]
\item How well-calibrated is \emph{explicit} metalinguistic output with respect to the system's \emph{implicit} linguistic generalizations? \citep{hu-23,song-25}
\item Can metalinguistic distinctions such as use vs.~mention be traced to internal model representations?
\item How much of model behavior on metalinguistic tasks can be attributed to metalinguistic text in pretraining data or data provided at inference time (or not, as discussed, e.g., in \citet{aycock-etal-2025-iclr} and \citet{marmonier-etal-2025-explicit})?
\item To the extent that metalinguistic meaning is grounded in linguistic usage, is distributional learning from form alone \citep[discussed, e.g., in][]{bender-20,pavlick-23} fundamentally different from learning of non-metalinguistic meaning?
\end{enumerate}

\paragraph{Extrinsic uses}
Finally, we delineate some under-explored uses of metalanguage for further downstream applications:
\begin{enumerate}[resume]
\item Can metalinguistic data such as syntax trees or grammar rules\slash descriptions be leveraged for inductive biases in pipelined systems \citep{wein-24}, integrated within LM architectures \citep{prange-22,gessler-23}, or via in-context learning \citep{court-elsner-2024-shortcomings,ginn-palmer-2025-llm,pei-etal-2025-understanding,nakashole-26,purushothama-coptic}?
\item How well can a system perform metalinguistic question answering? \citep{behzad-23} (discussed in \cref{sec:learning})
\item How can NLP shed light on how people use metalanguage? \citep{kranzlein-24} (discussed in \cref{sec:law})
\item Can we build LLM-powered assistants that deploy metalanguage effectively for language documentation, education, and scholarship?  (also discussed in \cref{sec:grammars} and \cref{sec:patterns})
\end{enumerate}

\section{Conclusion}

\iffalse
Points we want to make:
\begin{itemize}
    \item metalanguage is multi-faceted, metalinguistic abilities should be measured accordingly
    \item probing understanding via metalanguage is different from probing via next-word prediction or model-level interpretability; goal of LLM development should be to improve this calibration. Further, symbolic metalanguage has the confound of notation/framework knowledge; linguistic metalanguage has the confound of terminology: "syntactic raising" vs '"plain" raising'.
    \item  Learn generalization from metalanguage (learning from meta in general) and mapping it to usage. e.g. MT improve from side-information.
\end{itemize}
\fi

Metalanguage is inherently multifaceted. As we outlined in the possible dimensions in \cref{sec:discussion} above, it spans multiple levels of abstraction, heterogeneous representational formats, and diverse forms of linguistic reasoning. 
%As a result, no single benchmark could possibly adequately evaluate all necessary metalinguistic abilities of LLMs. 
This diversity should be taken into consideration as we 
%progress towards integrating metalanguage and associated abilites with LLMs.
devote greater attention to metalinguistic tasks and applications.

We are particularly excited about the interpretability questions around metalanguage.
%Probing models with metalanguage differs fundamentally from other paradigms such as mechanistic interpretability. 
%Unlike other approaches, 
Metalinguistic tasks may be particularly challenging where they require a model to \textit{both} articulate and apply linguistic reasoning. 

Metalanguage research also offers a promising pathway to studying learning and generalization. Humans frequently learn from explicit metalinguistic instructions and explanations and are able to apply this new knowledge to new examples.
One should expect models to be able to do the same. Advancing research on how models learn from and operationalize metalanguage should dramatically improve the frontier of LLM abilities in general.

\section*{Limitations}

We do not aim for this work to be a complete survey of metalanguage research. We by design draw heavily from our own work and our own perspective on the field.

\section*{Acknowledgments}

We thank our collaborators on our metalinguistic journeys, including members of the NERT lab and the George Mason NLP lab, Dr.~Amir Zeldes, and Dr.~Kevin Tobia.  
We benefited from the Metalinguistic NLP Bibliography (\url{https://github.com/nert-nlp/metalinguistic-nlp-bib}) spearheaded by Abhishek Purushothama.
This work was supported in part by NSF awards IIS-2144881 (Schneider) and IIS-2439202 (Anastasopoulos).

\bibliography{custom,nmlp}

@inproceedings{bender-20,
	address = {Online},
	title = {Climbing towards {NLU}: on meaning, form, and understanding in the age of data},
	url = {https://www.aclweb.org/anthology/2020.acl-main.463},
	booktitle = {Proc. of {ACL}},
	author = {Bender, Emily M. and Koller, Alexander},
	month = jul,
	year = {2020},
	pages = {5185--5198}
}

@article{pavlick-23,
	title = {Symbols and grounding in large language models},
	volume = {381},
	url = {https://royalsocietypublishing.org/doi/10.1098/rsta.2022.0041},
	number = {2251},
	journal = {Philosophical Transactions of the Royal Society A: Mathematical, Physical and Engineering Sciences},
	publisher = {Royal Society},
	author = {Pavlick, Ellie},
	month = jun,
	year = {2023},
	pages = {20220041}
}

@article{fillmore-85,
	title = {Frames and the semantics of understanding},
	volume = {6},
	url = {http://www.icsi.berkeley.edu/pubs/ai/framesand85.pdf},
	number = {2},
	journal = {Quaderni di Semantica},
	author = {Fillmore, Charles J.},
	year = {1985},
	pages = {222--254}
}

@article{horn-85,
	title = {Metalinguistic negation and pragmatic ambiguity},
	volume = {61},
	url = {https://www.jstor.org/stable/413423},
	number = {1},
	journal = {Language},
	publisher = {Linguistic Society of America},
	author = {Horn, Laurence R.},
	year = {1985},
	pages = {121--174}
}

@inproceedings{gessler-23,
	address = {Singapore},
	title = {Syntactic inductive bias in transformer language models: especially helpful for low-resource languages?},
	url = {https://aclanthology.org/2023.conll-1.17},
	booktitle = {Proc. of {CoNLL}},
	author = {Gessler, Luke and Schneider, Nathan},
	editor = {Jiang, Jing and Reitter, David and Deng, Shumin},
	month = dec,
	year = {2023},
	pages = {238--253}
}

@inproceedings{ettinger-23,
	address = {Singapore},
	title = {``{Y}ou are an expert linguistic annotator'': {L}imits of {LLMs} as analyzers of {A}bstract {M}eaning {R}epresentation},
	url = {https://aclanthology.org/2023.findings-emnlp.553/},
	booktitle = {Findings of the Association for Computational Linguistics: {EMNLP} 2023},
	author = {Ettinger, Allyson and Hwang, Jena and Pyatkin, Valentina and Bhagavatula, Chandra and Choi, Yejin},
	editor = {Bouamor, Houda and Pino, Juan and Bali, Kalika},
	month = dec,
	year = {2023},
	pages = {8250--8263}
}

@inproceedings{tian-24,
	address = {Bangkok, Thailand},
	title = {Large language models are no longer shallow parsers},
	url = {https://aclanthology.org/2024.acl-long.384},
	booktitle = {Proc. of {ACL}},
	author = {Tian, Yuanhe and Xia, Fei and Song, Yan},
	editor = {Ku, {Lun-Wei} and Martins, Andre and Srikumar, Vivek},
	month = aug,
	year = {2024},
	pages = {7131--7142}
}

@inproceedings{song-25,
	title = {Language models fail to introspect about their knowledge of language},
	url = {https://openreview.net/forum?id=AivRDOFi5H},
	booktitle = {Proc. of {COLM}},
	author = {Song, Siyuan and Hu, Jennifer and Mahowald, Kyle},
	month = aug,
	year = {2025}
}

@inproceedings{purushothama-facct,
	address = {Montréal, Canada},
	title = {Prompting from the bench: Large-scale pretraining is not sufficient to prepare {LLM}s for ordinary meaning analysis},
	booktitle = {Proc. of the Ninth Annual ACM Conference on Fairness, Accountability, and Transparency ({ACM} {FAccT})},
	author = {Purushothama, Abhishek  and  Min, Junghyun  and  Waldon, Brandon  and Schneider, Nathan},
	month = jun,
	year = {2026},
    url = {https://arxiv.org/abs/2510.25356},
    note = {{arXiv} preprint: 2510.25356 [cs]}
}

@inproceedings{petersen-26,
	address = {San Diego, California},
	title = {Sense and Sensitivity: ``Reasoning'' Models are More Robust, but can
Diverge from Human Consensus in a Legal Interpretation Task},
	booktitle = {Proc. of {CoNLL}},
	author = {Petersen, Dawson and Purushothama, Abhishek and Schneider, Nathan},
	month = jul,
	year = {2026}
}

@article{waldon-25-llms,
	title = {Large language models for legal interpretation? {D}on't take their word for it},
	volume = {114},
	url = {https://www.law.georgetown.edu/georgetown-law-journal/wp-content/uploads/sites/26/2026/02/Waldon_Schneider_Wilcox_Zeldes_Tobia_Large-Language-Models-for-Legal-Interpretation-Dont-Take-Their-Word-for-It.pdf},
	number = {1},
	journal = {Georgetown Law Journal},
	author = {Waldon, Brandon and Schneider, Nathan and Wilcox, Ethan and Zeldes, Amir and Tobia, Kevin},
	month = nov,
	year = {2025},
	pages = {115--183}
}

@phdthesis{kranzlein-diss,
	type = {{Ph.D.} dissertation},
	title = {Unpacking Meaning with Natural Language Processing: Legal Metalanguage Analysis and Long-Tail Calibration},
	url = {https://repository.library.georgetown.edu/handle/10822/1089042},
	school = {Georgetown University},
	author = {Kranzlein, Michael},
	year = {2024}
}

@inproceedings{prange-22,
	address = {Seattle, United States},
	title = {Linguistic frameworks go toe-to-toe at neuro-symbolic language modeling},
	url = {https://aclanthology.org/2022.naacl-main.325},
	booktitle = {Proc. of {NAACL-HLT}},
	author = {Prange, Jakob and Schneider, Nathan and Kong, Lingpeng},
	month = jul,
	year = {2022},
	pages = {4375--4391}
}

@inproceedings{wein-24,
	address = {St. Julian's, Malta},
	title = {Lost in translationese? {R}educing translation effect using {A}bstract {M}eaning {R}epresentation},
	url = {https://aclanthology.org/2024.eacl-long.45},
	booktitle = {Proc. of {EACL}},
	author = {Wein, Shira and Schneider, Nathan},
	editor = {Graham, Yvette and Purver, Matthew},
	month = mar,
	year = {2024},
	pages = {753--765}
}

@inproceedings{gligoric-24,
	address = {Mexico City, Mexico},
	title = {{NLP} systems that can't tell use from mention censor counterspeech, but teaching the distinction helps},
	url = {https://aclanthology.org/2024.naacl-long.331/},
	booktitle = {Proc. of {NAACL-HLT}},
	author = {Gligorić, Kristina and Cheng, Myra and Zheng, Lucia and Durmus, Esin and Jurafsky, Dan},
	editor = {Duh, Kevin and Gomez, Helena and Bethard, Steven},
	month = jun,
	year = {2024},
	pages = {5942--5959}
}

@phdthesis{wilson-11,
	address = {College Park, Maryland},
	type = {{Ph.D.} dissertation},
	title = {A computational theory of the use-mention distinction in natural language},
	url = {https://www.proquest.com/docview/881106306/abstract/E4FCF769753E446DPQ/1},
	school = {University of Maryland},
	author = {Wilson, Shomir},
	year = {2011}
}

@inproceedings{aycock-etal-2025-iclr,
 author = {Aycock, Seth and Stap, David and Wu, Di and Monz, Christof and Simaan, Khalil},
 booktitle = {International Conference on Learning Representations},
 editor = {Y. Yue and A. Garg and N. Peng and F. Sha and R. Yu},
 pages = {12334--12357},
 title = {Can {LLM}s Really Learn to Translate a Low-Resource Language from One Grammar Book?},
 url = {https://proceedings.iclr.cc/paper_files/paper/2025/file/20f44da80080d76bbc35bca0027f14e6-Paper-Conference.pdf},
 volume = {2025},
 year = {2025}
}

@inproceedings{marmonier-etal-2025-explicit,
    title = "Explicit Learning and the {LLM} in Machine Translation",
    author = "Marmonier, Malik  and
      Bawden, Rachel  and
      Sagot, Beno{\^i}t",
    editor = "Christodoulopoulos, Christos  and
      Chakraborty, Tanmoy  and
      Rose, Carolyn  and
      Peng, Violet",
    booktitle = "Proceedings of the 2025 Conference on Empirical Methods in Natural Language Processing",
    month = nov,
    year = "2025",
    address = "Suzhou, China",
    publisher = "Association for Computational Linguistics",
    url = "https://aclanthology.org/2025.emnlp-main.1599/",
    doi = "10.18653/v1/2025.emnlp-main.1599",
    pages = "31372--31422",
    ISBN = "979-8-89176-332-6"
}

@inproceedings{hus-anastasopoulos-2024-back,
    title = "Back to School: Translation Using Grammar Books",
    author = "Hus, Jonathan  and
      Anastasopoulos, Antonios",
    editor = "Al-Onaizan, Yaser  and
      Bansal, Mohit  and
      Chen, Yun-Nung",
    booktitle = "Proceedings of the 2024 Conference on Empirical Methods in Natural Language Processing",
    month = nov,
    year = "2024",
    address = "Miami, Florida, USA",
    publisher = "Association for Computational Linguistics",
    url = "https://aclanthology.org/2024.emnlp-main.1127/",
    doi = "10.18653/v1/2024.emnlp-main.1127",
    pages = "20207--20219"
}

@inproceedings{hus-anastasopoulos-2026-sigtyp,
    title = "A RAG Approach for Typological Database Completion",
    author = "Hus, Jonathan  and
      Anastasopoulos, Antonios",
    booktitle = "Proceedings of the Eighth Workshop on Computational Research in Linguistic Typology",
    month = apr,
    year = "2026",
    address = "Rabbat, Morocco",
    publisher = "Association for Computational Linguistics",
}

@inproceedings{taguchi2025digital,
  title={Digital Documentation for Diasporic Data: challenges, opportunities, and solutions for working with Diaspora Communities},
  author={Taguchi, Chihiro and Liebl, J Elizabeth and Anastasopoulos, Antonios and Chiang, David and Walther, G{\'e}raldine},
  year={2025},
  booktitle={9th International Conference on Language Documentation \& Conservation (ICLDC)}
}

@inproceedings{tanzer2023mtob,
 author = {Tanzer, Garrett and Suzgun, Mirac and Visser, Eline and Jurafsky, Dan and Melas-Kyriazi, Luke},
 booktitle = {International Conference on Learning Representations},
 editor = {B. Kim and Y. Yue and S. Chaudhuri and K. Fragkiadaki and M. Khan and Y. Sun},
 pages = {18955--18985},
 title = {A Benchmark for Learning to Translate a New Language from One Grammar Book},
 url = {https://proceedings.iclr.cc/paper_files/paper/2024/file/52d63f9e4b81f866bf69fb3c834aad47-Paper-Conference.pdf},
 volume = {2024},
 year = {2024}
}

@inproceedings{bird-2022-local,
    title = "Local Languages, Third Spaces, and other High-Resource Scenarios",
    author = "Bird, Steven",
    editor = "Muresan, Smaranda  and
      Nakov, Preslav  and
      Villavicencio, Aline",
    booktitle = "Proceedings of the 60th Annual Meeting of the Association for Computational Linguistics (Volume 1: Long Papers)",
    month = may,
    year = "2022",
    address = "Dublin, Ireland",
    publisher = "Association for Computational Linguistics",
    url = "https://aclanthology.org/2022.acl-long.539/",
    doi = "10.18653/v1/2022.acl-long.539",
    pages = "7817--7829"
}

@inproceedings{chaudhary-etal-2023-teacher,
    title = "Teacher Perception of Automatically Extracted Grammar Concepts for {L}2 Language Learning",
    author = "Chaudhary, Aditi  and
      Sampath, Arun  and
      Sheshadri, Ashwin  and
      Anastasopoulos, Antonios  and
      Neubig, Graham",
    editor = "Bouamor, Houda  and
      Pino, Juan  and
      Bali, Kalika",
    booktitle = "Findings of the Association for Computational Linguistics: EMNLP 2023",
    month = dec,
    year = "2023",
    address = "Singapore",
    publisher = "Association for Computational Linguistics",
    url = "https://aclanthology.org/2023.findings-emnlp.246/",
    doi = "10.18653/v1/2023.findings-emnlp.246",
    pages = "3776--3793"
}

@inproceedings{chaudhary-etal-2021-wall,
    title = "When is \emph{wall} a \emph{pared} and when a \emph{muro}?: Extracting rules governing lexical selection",
    author = "Chaudhary, Aditi  and
      Yin, Kayo  and
      Anastasopoulos, Antonios  and
      Neubig, Graham",
    editor = "Moens, Marie-Francine  and
      Huang, Xuanjing  and
      Specia, Lucia  and
      Yih, Scott Wen-tau",
    booktitle = "Proceedings of the 2021 Conference on Empirical Methods in Natural Language Processing",
    month = nov,
    year = "2021",
    address = "Online and Punta Cana, Dominican Republic",
    publisher = "Association for Computational Linguistics",
    url = "https://aclanthology.org/2021.emnlp-main.553/",
    doi = "10.18653/v1/2021.emnlp-main.553",
    pages = "6911--6929"
}

@inproceedings{daume-07,
	address = {Prague, Czech Republic},
	title = {A {B}ayesian model for discovering typological implications},
	url = {http://www.aclweb.org/anthology/P07-1009},
	booktitle = {Proc. of {ACL}},
	author = {Daum\'{e}, {III}, Hal and Campbell, Lyle},
	month = jun,
	year = {2007},
	pages = {65--72}
}

@article{grambank,
	title = {Grambank reveals the importance of genealogical constraints on linguistic diversity and highlights the impact of language loss},
	volume = {9},
	doi = {https://doi.org/10.1126/sciadv.adg6175},
	number = {16},
	journal = {Science Advances},
	publisher = {American Association for the Advancement of Science},
	author = {Skirg\r{a}rd, Hedvig and Haynie, Hannah J. and Blasi, Dami\'{a}n E. and Hammarstr\"{o}m, Harald and Collins, Jeremy and Latarche, Jay J. and Lesage, Jakob and Weber, Tobias and {Witzlack-Makarevich}, Alena and Passmore, Sam and Chira, Angela and Maurits, Luke and Dinnage, Russell and Dunn, Michael and Reesink, Ger and Singer, Ruth and Bowern, Claire and Epps, Patience and Hill, Jane and Vesakoski, Outi and Robbeets, Martine and Abbas, Noor Karolin and Auer, Daniel and Bakker, Nancy A. and Barbos, Giulia and Borges, Robert D. and Danielsen, Swintha and Dorenbusch, Luise and Dorn, Ella and Elliott, John and Falcone, Giada and Fischer, Jana and Ghanggo Ate, Yustinus and Gibson, Hannah and G\"{o}bel, {Hans-Philipp} and Goodall, Jemima A. and Gruner, Victoria and Harvey, Andrew and Hayes, Rebekah and Heer, Leonard and Herrera Miranda, Roberto E. and H\"{u}bler, Nataliia and {Huntington-Rainey}, Biu and Ivani, Jessica K. and Johns, Marilen and Just, Erika and Kashima, Eri and Kipf, Carolina and Klingenberg, Janina V. and K\"{o}nig, Nikita and Koti, Aikaterina and Kowalik, Richard G.~A. and Krasnoukhova, Olga and Lindvall, Nora L.~M. and Lorenzen, Mandy and Lutzenberger, Hannah and Martins, T\^{a}nia R.~A. and Mata German, Celia and van der Meer, Suzanne and Montoya Samam\'{e}, Jaime and M\"{u}ller, Michael and Muradoglu, Saliha and Neely, Kelsey and Nickel, Johanna and Norvik, Miina and Oluoch, Cheryl Akinyi and Peacock, Jesse and Pearey, India O.~C. and Peck, Naomi and Petit, Stephanie and Pieper, S\"{o}ren and Poblete, Mariana and Prestipino, Daniel and Raabe, Linda and Raja, Amna and Reimringer, Janis and Rey, Sydney C. and Rizaew, Julia and Ruppert, Eloisa and Salmon, Kim K. and Sammet, Jill and Schembri, Rhiannon and Schlabbach, Lars and Schmidt, Frederick W.~P. and Skilton, Amalia and Smith, Wikaliler Daniel and de Sousa, Hil\'{a}rio and Sverredal, Kristin and Valle, Daniel and Vera, Javier and Vo{\ss}, Judith and Witte, Tim and Wu, Henry and Yam, Stephanie and Ye, Jingting and Yong, Maisie and Yuditha, Tessa and Zariquiey, Roberto and Forkel, Robert and Evans, Nicholas and Levinson, Stephen C. and Haspelmath, Martin and Greenhill, Simon J. and Atkinson, Quentin D. and Gray, Russell D.},
	month = apr,
	year = {2023},
	pages = {eadg6175}
}

@misc{snell,
  title = {Concurring opinion in \textit{Snell v. United Specialty Insurance Co.}},
author={Newsom, Kevin},
  note = {United States Court of Appeals
For the Eleventh Circuit, 22-12581},
  year = {2024},
  url = {https://media.ca11.uscourts.gov/opinions/pub/files/202212581.pdf} 
}

@book{wals,
  address   = {Leipzig},
  editor    = {Matthew S. Dryer and Martin Haspelmath},
  publisher = {Max Planck Institute for Evolutionary Anthropology},
  title     = {WALS Online},
  url       = {https://wals.info/},
  year      = {2013}
}

@book{harris1951methods,
  title={Methods in Structural Linguistics},
  author={Harris, Zellig S.},
  year={1951},
  publisher={University of Chicago Press}
}

@inproceedings{howell2017inferring,
  title={Inferring case systems from IGT: Enriching the enrichment},
  author={Howell, Kristen and Bender, Emily M and Lockwood, Michel and Xia, Fei and Zamaraeva, Olga},
  booktitle={Proceedings of the 2nd Workshop on the Use of Computational Methods in the Study of Endangered Languages},
  pages={67--75},
  year={2017}
}

@inproceedings{zamaraeva2016inferring,
    title = "Inferring Morphotactics from Interlinear Glossed Text: Combining Clustering and Precision Grammars",
    author = "Zamaraeva, Olga",
    editor = "Elsner, Micha  and
      Kuebler, Sandra",
    booktitle = "Proc. of the {SIGMORPHON} Workshop on Computational Research in Phonetics, Phonology, and Morphology",
    month = aug,
    year = "2016",
    address = "Berlin, Germany",
    publisher = "Association for Computational Linguistics",
    url = "https://aclanthology.org/W16-2021/",
    doi = "10.18653/v1/W16-2021",
    pages = "141--150"
}

@inproceedings{baylor-etal-2023-past,
    title = "The Past, Present, and Future of Typological Databases in {NLP}",
    author = "Baylor, Emi  and
      Ploeger, Esther  and
      Bjerva, Johannes",
    editor = "Bouamor, Houda  and
      Pino, Juan  and
      Bali, Kalika",
    booktitle = "Findings of the Association for Computational Linguistics: EMNLP 2023",
    month = dec,
    year = "2023",
    address = "Singapore",
    publisher = "Association for Computational Linguistics",
    url = "https://aclanthology.org/2023.findings-emnlp.82/",
    doi = "10.18653/v1/2023.findings-emnlp.82",
    pages = "1163--1169"
}

@inproceedings{bjerva-etal-2020-sigtyp,
    title = "{SIGTYP} 2020 Shared Task: Prediction of Typological Features",
    author = "Bjerva, Johannes  and
      Salesky, Elizabeth  and
      Mielke, Sabrina J.  and
      Chaudhary, Aditi  and
      Celano, Giuseppe G. A.  and
      Ponti, Edoardo Maria  and
      Vylomova, Ekaterina  and
      Cotterell, Ryan  and
      Augenstein, Isabelle",
    editor = "Vylomova, Ekaterina  and
      Ponti, Edoardo M.  and
      Grossman, Eitan  and
      McCarthy, Arya D.  and
      Berzak, Yevgeni  and
      Dubossarsky, Haim  and
      Vuli{\'c}, Ivan  and
      Reichart, Roi  and
      Korhonen, Anna  and
      Cotterell, Ryan",
    booktitle = "Proceedings of the Second Workshop on Computational Research in Linguistic Typology",
    month = nov,
    year = "2020",
    address = "Online",
    publisher = "Association for Computational Linguistics",
    url = "https://aclanthology.org/2020.sigtyp-1.1/",
    doi = "10.18653/v1/2020.sigtyp-1.1",
    pages = "1--11"
}

@inproceedings{sanchez-25,
	title = {Linguini: {A} benchmark for language-agnostic linguistic reasoning},
	url = {https://openreview.net/forum?id=CCBPSxWOhi},
	booktitle = {The Thirty-ninth Annual Conference on Neural Information Processing Systems Datasets and Benchmarks Track},
	author = {S\'{a}nchez, Eduardo and Alastruey, Belen and Ropers, Christophe and Turkatenko, Arina and Stenetorp, Pontus and Artetxe, Mikel and Costa-juss\`{a}, Marta R.},
	month = oct,
	year = {2025}
}

@misc{bean2024lingoly,
	title = {{LINGOLY}: {A} benchmark of olympiad-level linguistic reasoning puzzles in low-resource and extinct languages},
	volume = {37},
	url = {https://proceedings.neurips.cc/paper_files/paper/2024/hash/2e43584b7d7b32fb6b2aa83b32dbbb20-Abstract-Datasets_and_Benchmarks_Track.html},
	journal = {Advances in Neural Information Processing Systems},
	author = {Bean, Andrew and Hellsten, Simi and Mayne, Harry and Magomere, Jabez and A., Ethan and Chi, Ryan and Hale, Scott A. and Kirk, Hannah R.},
	month = dec,
	year = {2024},
	pages = {26224--26237}
}

@inproceedings{yang2025linggym,
	address = {Suzhou, China},
	title = {{LingGym}: {H}ow far are {LLMs} from thinking like field linguists?},
	url = {https://aclanthology.org/2025.emnlp-main.69/},
	booktitle = {Proc. of {EMNLP}},
	author = {Yang, Changbing and Ma, Franklin and Shi, Freda and Zhu, Jian},
	editor = {Christodoulopoulos, Christos and Chakraborty, Tanmoy and Rose, Carolyn and Peng, Violet},
	month = nov,
	year = {2025},
	pages = {1314--1340}
}

@techreport{brannon-23-interp,
	type = {Report},
	title = {Statutory interpretation: theories, tools, and trends},
	url = {https://www.congress.gov/crs-product/R45153},
	number = {R45153},
	institution = {Congressional Research Service},
	author = {Brannon, Valerie C.},
	month = mar,
	year = {2023}
}

@techreport{brannon-25-canons,
	type = {In Focus},
	title = {Canons of construction: a brief overview},
	url = {https://www.congress.gov/crs-product/IF12992},
	number = {{IF12992}},
	institution = {Congressional Research Service},
	author = {Brannon, Valerie C.},
	month = may,
	year = {2025}
}

@book{scalia-12,
	address = {St. Paul, {MN}},
	title = {Reading law: the interpretation of legal texts},
	publisher = {Thomson/West},
	author = {Scalia, Antonin and Garner, Bryan A.},
	year = {2012}
}

@article{eskridge-23,
	title = {Textualism's defining moment},
	volume = {123},
	url = {https://www.jstor.org/stable/27259434},
	number = {6},
	journal = {Columbia Law Review},
	publisher = {{JSTOR}},
	author = {Eskridge, William N. and Slocum, Brian G. and Tobia, Kevin},
	year = {2023},
	pages = {1611--1698}
}

@article{tobia-20,
	title = {Testing ordinary meaning},
	volume = {134},
	url = {https://www.jstor.org/stable/27028333},
	number = {2},
	journal = {Harvard Law Review},
	author = {Tobia, Kevin P.},
	month = dec,
	year = {2020},
	pages = {726--806}
}

@article{tobia-22,
	title = {Experimental jurisprudence},
	volume = {89},
	url = {https://www.jstor.org/stable/27132267},
	number = {3},
	journal = {The University of Chicago Law Review},
	publisher = {{JSTOR}},
	author = {Tobia, Kevin},
	year = {2022},
	pages = {735--802}
}

@article{waldon-25-artifact,
	title = {Reading law with linguistics: the statutory interpretation of artifact nouns},
	volume = {62},
	url = {https://journals.law.harvard.edu/jol/2025/06/01/tobia-linguistics/},
	number = {2},
	journal = {Harvard Journal on Legislation},
	author = {Waldon, Brandon and Condoravdi, Cleo and Pustejovsky, James and Schneider, Nathan and Tobia, Kevin},
	month = jun,
	year = {2025},
	pages = {415--467}
}

@article{tobia-22-canons,
	title = {Statutory {I}nterpretation from the outside},
	volume = {122},
	url = {https://heinonline.org/HOL/P?h=hein.journals/clr122&i=221},
	number = {1},
	journal = {Columbia Law Review},
	author = {Tobia, Kevin and Slocum, Brian G. and Nourse, Victoria},
	year = {2022},
	pages = {213--330}
}

@inproceedings{waldon-25-legalcgel,
	address = {Ljubljana, Slovenia},
	title = {{Legal-CGEL}: {A}nalyzing legal text in the {CGELBank} framework},
	url = {https://aclanthology.org/2025.tlt-1.17/},
	booktitle = {Proc. of the 23rd International Workshop on Treebanks and Linguistic Theories ({TLT}, {SyntaxFest} 2025)},
	author = {Waldon, Brandon and Wells, Micaela and Tiwari, Devika and Gopalan, Meru and Schneider, Nathan},
	editor = {Jablotschkin, Sarah and K\"{u}bler, Sandra and Zinsmeister, Heike},
	month = aug,
	year = {2025},
	pages = {148--153}
}

@book{breyer-23,
	title = {County of {M}aui v. {H}awaii {W}ildlife {F}und},
	note = {140 S. Ct. 1462},
	url = {https://supreme.justia.com/cases/federal/us/590/18-260/case.pdf},
	author = {Breyer, Stephen},
	month = apr,
	year = {2023}
}

@article{opitz-25,
	title = {Natural language processing {RELIES} on linguistics},
	volume = {51},
	url = {https://doi.org/10.1162/coli_a_00560},
	number = {3},
	journal = {Computational Linguistics},
	author = {Opitz, Juri and Wein, Shira and Schneider, Nathan},
	month = sep,
	year = {2025},
	pages = {1009--1032}
}

@inproceedings{liu-19,
	address = {Minneapolis, Minnesota},
	title = {Linguistic knowledge and transferability of contextual representations},
	url = {https://www.aclweb.org/anthology/N19-1112},
	booktitle = {Proc. of {NAACL-HLT}},
	author = {Liu, Nelson F. and Gardner, Matt and Belinkov, Yonatan and Peters, Matthew E. and Smith, Noah A.},
	month = jun,
	year = {2019},
	pages = {1073--1094}
}

@inproceedings{tenney-19,
	address = {Florence, Italy},
	title = {{BERT} rediscovers the classical {NLP} pipeline},
	url = {https://aclanthology.org/P19-1452},
	booktitle = {Proc. of {ACL}},
	author = {Tenney, Ian and Das, Dipanjan and Pavlick, Ellie},
	month = jul,
	year = {2019},
	pages = {4593--4601}
}

@inproceedings{aoyama-22,
	address = {Hybrid: Seattle, Washington + Online},
	title = {Probe-less probing of {BERT's} layer-wise linguistic knowledge with masked word prediction},
	url = {https://aclanthology.org/2022.naacl-srw.25},
	booktitle = {Proc. of the 2022 Conference of the North American Chapter of the Association for Computational Linguistics: Human Language Technologies: Student Research Workshop},
	author = {Aoyama, Tatsuya and Schneider, Nathan},
	month = jul,
	year = {2022},
	pages = {195--201}
}

@article{blimp,
	title = {{BLiMP}: {T}he {B}enchmark of {L}inguistic {M}inimal {P}airs for {E}nglish},
	volume = {8},
	url = {https://doi.org/10.1162/tacl_a_00321},
	journal = {Transactions of the Association for Computational Linguistics},
	publisher = {{MIT} Press},
	author = {Warstadt, Alex and Parrish, Alicia and Liu, Haokun and Mohananey, Anhad and Peng, Wei and Wang, {Sheng-Fu} and Bowman, Samuel R.},
	month = jul,
	year = {2020},
	pages = {377--392}
}

@misc{wang-22,
	title = {Interpretability in the wild: a circuit for indirect object identification in {GPT-2} small},
	url = {http://arxiv.org/abs/2211.00593},
	publisher = {{arXiv}},
	author = {Wang, Kevin and Variengien, Alexandre and Conmy, Arthur and Shlegeris, Buck and Steinhardt, Jacob},
	month = nov,
	year = {2022},
	note = {{arXiv}:2211.00593 [cs]}
}

@inproceedings{purushothama-coptic,
	title = {Syntax as a {R}osetta {S}tone: {U}niversal {D}ependencies for In-Context {C}optic Translation},
    booktitle = {Findings of ACL},
	url = {http://arxiv.org/abs/2604.18758},
	author = {Purushothama, Abhishek and Thronson, Emma and Guo, Alexia and Zeldes, Amir},
	month = jul,
	year = {2026},
	note = {{arXiv} preprint: 2604.18758 [cs]}
}

@inproceedings{pei-etal-2025-understanding,
    title = "Understanding In-Context Machine Translation for Low-Resource Languages: A Case Study on {M}anchu",
    author = "Pei, Renhao  and
      Liu, Yihong  and
      Lin, Peiqin  and
      Yvon, Fran{\c{c}}ois  and
      Schuetze, Hinrich",
    editor = "Che, Wanxiang  and
      Nabende, Joyce  and
      Shutova, Ekaterina  and
      Pilehvar, Mohammad Taher",
    booktitle = "Proceedings of the 63rd Annual Meeting of the Association for Computational Linguistics (Volume 1: Long Papers)",
    month = jul,
    year = "2025",
    address = "Vienna, Austria",
    publisher = "Association for Computational Linguistics",
    url = "https://aclanthology.org/2025.acl-long.429/",
    doi = "10.18653/v1/2025.acl-long.429",
    pages = "8767--8788",
    ISBN = "979-8-89176-251-0"
}

@inproceedings{court-elsner-2024-shortcomings,
    title = "Shortcomings of {LLM}s for Low-Resource Translation: Retrieval and Understanding Are Both the Problem",
    author = "Court, Sara  and
      Elsner, Micha",
    editor = "Haddow, Barry  and
      Kocmi, Tom  and
      Koehn, Philipp  and
      Monz, Christof",
    booktitle = "Proceedings of the Ninth Conference on Machine Translation",
    month = nov,
    year = "2024",
    address = "Miami, Florida, USA",
    publisher = "Association for Computational Linguistics",
    url = "https://aclanthology.org/2024.wmt-1.125/",
    doi = "10.18653/v1/2024.wmt-1.125",
    pages = "1332--1354"
}

@inproceedings{nakashole-26,
    author = {Nakashole, Ndapa},
    title = {Grammar as Control: Modular Language Generation for the Long Tail},
    booktitle = {Proc. of ACL},
    year = {2026},
    month = jul,
    url = {https://ndapa.us/assets/docs/papers/2026-acl-mtig.pdf},
    address = {San Diego, California}
}

@inproceedings{wells-25-abstract,
	title = {Scope ambiguity resolution of negated connectives in {E}nglish corpora},
	volume = {47},
	url = {https://escholarship.org/uc/item/3xp1608t},
	booktitle = {Proc. of the Annual Meeting of the Cognitive Science Society},
	author = {Wells, Micaela and Waldon, Brandon and Schneider, Nathan},
	year = {2025},
	pages = {6608}
}

@inproceedings{ginn-palmer-2025-llm,
    title = "{LLM} Dependency Parsing with In-Context Rules",
    author = "Ginn, Michael  and
      Palmer, Alexis",
    editor = "Fei, Hao  and
      Tu, Kewei  and
      Zhang, Yuhui  and
      Hu, Xiang  and
      Han, Wenjuan  and
      Jia, Zixia  and
      Zheng, Zilong  and
      Cao, Yixin  and
      Zhang, Meishan  and
      Lu, Wei  and
      Siddharth, N.  and
      {\O}vrelid, Lilja  and
      Xue, Nianwen  and
      Zhang, Yue",
    booktitle = "Proceedings of the 1st Joint Workshop on Large Language Models and Structure Modeling (XLLM 2025)",
    month = aug,
    year = "2025",
    address = "Vienna, Austria",
    publisher = "Association for Computational Linguistics",
    url = "https://aclanthology.org/2025.xllm-1.17/",
    doi = "10.18653/v1/2025.xllm-1.17",
    pages = "186--196",
    ISBN = "979-8-89176-286-2"
}

@inproceedings{wilson-13,
	title = {Toward automatic processing of {E}nglish metalanguage},
	url = {https://www.aclweb.org/anthology/I13-1091},
	booktitle = {Proc. of {IJCNLP}},
	author = {Wilson, Shomir},
	year = {2013}
}

@inproceedings{sahin-20,
	title = {{PuzzLing} {M}achines: a challenge on learning from small data},
	url = {https://www.aclweb.org/anthology/2020.acl-main.115},
	booktitle = {Proc. of {ACL}},
	author = {{\c S}ahin, Gözde Gül and Kementchedjhieva, Yova and Rust, Phillip and Gurevych, Iryna},
	year = {2020}
}

@inproceedings{rozner-21,
	title = {Decrypting cryptic crosswords: semantically complex wordplay puzzles as a target for {NLP}},
	volume = {34},
	url = {https://proceedings.neurips.cc/paper/2021/hash/5f1d3986fae10ed2994d14ecd89892d7-Abstract.html},
	booktitle = {Advances in Neural Information Processing Systems},
	author = {Rozner, Josh and Potts, Christopher and Mahowald, Kyle},
	year = {2021},
	pages = {11409--11421}
}

@article{berry-05,
	title = {Making the most of metalanguage},
	volume = {14},
	url = {http://www.tandfonline.com/doi/abs/10.1080/09658410508668817},
	number = {1},
	journal = {Language Awareness},
	author = {Berry, Roger},
	month = feb,
	year = {2005},
	pages = {3--20}
}

@article{begus-25,
	title = {Large linguistic models: investigating {LLMs}' metalinguistic abilities},
	volume = {6},
	url = {https://ieeexplore.ieee.org/abstract/document/11022724},
	number = {12},
	journal = {{IEEE} Transactions on Artificial Intelligence},
	author = {Begu\v{s}, Ga\v{s}per and D\k{a}bkowski, Maksymilian and Rhodes, Ryan},
	month = dec,
	year = {2025},
	pages = {3453--3467}
}

@inproceedings{behzad-23,
	title = {{ELQA}: {A} corpus of metalinguistic questions and answers about {E}nglish},
	url = {https://aclanthology.org/2023.acl-long.113},
	booktitle = {Proc. of {ACL}},
	author = {Behzad, Shabnam and Sakaguchi, Keisuke and Schneider, Nathan and Zeldes, Amir},
	year = {2023}
}

@inproceedings{kranzlein-24,
	title = {{CuRIAM}: {C}orpus {R}e {I}nterpretation and {M}etalanguage in {U.{S}.} {S}upreme {C}ourt {O}pinions},
	url = {https://aclanthology.org/2024.lrec-main.379},
	booktitle = {Proc. of {LREC-COLING}},
	author = {Kranzlein, Michael and Schneider, Nathan and Tobia, Kevin},
	editor = {Calzolari, Nicoletta and Kan, {Min-Yen} and Hoste, Veronique and Lenci, Alessandro and Sakti, Sakriani and Xue, Nianwen},
	year = {2024}
}

@inproceedings{hu-23,
	title = {Prompting is not a substitute for probability measurements in large language models},
	url = {https://aclanthology.org/2023.emnlp-main.306},
	booktitle = {Proc. of {EMNLP}},
	author = {Hu, Jennifer and Levy, Roger},
	editor = {Bouamor, Houda and Pino, Juan and Bali, Kalika},
	year = {2023}
}

@inproceedings{chaudhary-20,
	title = {Automatic extraction of rules governing morphological agreement},
	url = {https://aclanthology.org/2020.emnlp-main.422},
	booktitle = {Proc. of {EMNLP}},
	author = {Chaudhary, Aditi and Anastasopoulos, Antonios and Pratapa, Adithya and Mortensen, David R. and Sheikh, Zaid and Tsvetkov, Yulia and Neubig, Graham},
	editor = {Webber, Bonnie and Cohn, Trevor and He, Yulan and Liu, Yang},
	year = {2020}
}

@inproceedings{thrush-24,
	title = {I am a strange dataset: metalinguistic tests for language models},
	url = {https://aclanthology.org/2024.acl-long.482},
	booktitle = {Proc. of {ACL}},
	author = {Thrush, Tristan and Moore, Jared and Monares, Miguel and Potts, Christopher and Kiela, Douwe},
	editor = {Ku, {Lun-Wei} and Martins, Andre and Srikumar, Vivek},
	year = {2024}
}

@inproceedings{chi-24,
	title = {{ModeLing}: a novel dataset for testing linguistic reasoning in language models},
	url = {https://aclanthology.org/2024.sigtyp-1.14/},
	booktitle = {Proc. of the 6th Workshop on Research in Computational Linguistic Typology and Multilingual {NLP}},
	author = {Chi, Nathan and Malchev, Teodor and Kong, Riley and Chi, Ryan and Huang, Lucas and Chi, Ethan and {McCoy}, R. and Radev, Dragomir},
	editor = {Hahn, Michael and Sorokin, Alexey and Kumar, Ritesh and Shcherbakov, Andreas and Otmakhova, Yulia and Yang, Jinrui and Serikov, Oleg and Rani, Priya and Ponti, Edoardo M. and Muradoğlu, Saliha and Gao, Rena and Cotterell, Ryan and Vylomova, Ekaterina},
	year = {2024}
}

@inproceedings{behzad-etal-2024-ask,
    title = "To Ask {LLM}s about {E}nglish Grammaticality, Prompt Them in a Different Language",
    author = "Behzad, Shabnam  and
      Zeldes, Amir  and
      Schneider, Nathan",
    editor = "Al-Onaizan, Yaser  and
      Bansal, Mohit  and
      Chen, Yun-Nung",
    booktitle = "Findings of the Association for Computational Linguistics: EMNLP 2024",
    month = nov,
    year = "2024",
    address = "Miami, Florida, USA",
    publisher = "Association for Computational Linguistics",
    url = "https://aclanthology.org/2024.findings-emnlp.916/",
    doi = "10.18653/v1/2024.findings-emnlp.916",
    pages = "15622--15634"
}

@inproceedings{choudhary2025unveiling,
	title        = {{UNVEILING}: What Makes Linguistics Olympiad Puzzles Tricky for {LLM}s?},
	author       = {Mukund Choudhary and KV Aditya Srivatsa and Gaurja Aeron and Antara Raaghavi Bhattacharya and Dang Khoa Dang Dinh and Ikhlasul Akmal Hanif and Daria Kotova and Ekaterina Kochmar and Monojit Choudhury},
	year         = 2025,
	booktitle    = {Proc. of {COLM}},
	url          = {https://openreview.net/forum?id=fcRcl1EXc4}
}

@misc{arcon-26,
	title = {Evaluating Metalinguistic Knowledge in Large Language Models across the World's Languages},
	url = {http://arxiv.org/abs/2602.02182},
	publisher = {{arXiv}},
	author = {Ar\v{c}on, Tja\v{s}a and Klemen, Matej and Robnik-\v{S}ikonja, Marko and Dobrovoljc, Kaja},
	month = feb,
	year = {2026},
	note = {{arXiv}:2602.02182 [cs]}
}

\iffalse
\appendix

\section{Example Appendix}
\label{sec:appendix}

This is an appendix.
\fi

\end{document}